\documentclass[conference]{IEEEtran}
\IEEEoverridecommandlockouts
\usepackage{cite}
\usepackage{amsmath,amssymb,amsfonts}
\usepackage{algorithmic}
\usepackage{graphicx}
\usepackage{textcomp}
\usepackage{xcolor}
\usepackage{xspace}
\usepackage{url}
\usepackage{booktabs}  
\usepackage{enumitem}
\usepackage{subcaption} 
\usepackage{soul}
\def\BibTeX{{\rm B\kern-.05em{\sc i\kern-.025em b}\kern-.08em
    T\kern-.1667em\lower.7ex\hbox{E}\kern-.125emX}}
\begin{document}

\title{
SplitSEE: A Splittable Self-supervised Framework for Single-Channel EEG Representation Learning
}
\author{
    \IEEEauthorblockN{Rikuto Kotoge\IEEEauthorrefmark{1}\IEEEauthorrefmark{4}, Zheng Chen\IEEEauthorrefmark{1}\IEEEauthorrefmark{4}, Tasuku Kimura\IEEEauthorrefmark{1}, Yasuko Matsubara\IEEEauthorrefmark{1}, \\
    Takufumi Yanagisawa\IEEEauthorrefmark{2}\IEEEauthorrefmark{3}, Haruhiko Kishima\IEEEauthorrefmark{2}, Yasushi Sakurai\IEEEauthorrefmark{1}}
    \IEEEauthorblockA{\IEEEauthorrefmark{1}SANKEN, Osaka University, Osaka, Japan}
    \IEEEauthorblockA{\IEEEauthorrefmark{2}Department of Neurosurgery, Graduate School of Medicine, Osaka University, Osaka, Japan}
    \IEEEauthorblockA{\IEEEauthorrefmark{3}Institute for Advanced Co-Creation Studies, Osaka University, Osaka, Japan \\\{rikuto88, chenz, tasuku, yasuko, yasushi\}@sanken.osaka-u.ac.jp \\\{tyanagisawa, hkishima\}@nsurg.med.osaka-u.ac.jp}
    \thanks{\IEEEauthorrefmark{4} indicates corresponding authors.}
}

\maketitle

\newcommand{\goal}[1]{ {\noindent {$\Rightarrow$} \em {#1} } }
\newcommand{\hide}[1]{}
\newcommand{\reminder}[1]{{\textsf{\textcolor{red}{[#1]}}}}
\newcommand{\bluereminder}[1]{{\textsf{\textcolor{blue}{[#1]}}}}
\newcommand{\duty}[1]{{\textsf{\textcolor{blue}{[#1's job]}}}}
\newcommand{\vectornorm}[1]{\left|\left|#1\right|\right|}
\newcommand{\mypara}[1]{\vspace{0.40em}\noindent{\bf #1.}}
\newcommand{\myparad}[1]{\vspace{0.40em}\noindent{\bf #1}}
\newcommand{\myipara}[1]{\vspace{0.40em}\noindent{\em #1:}}

\newcommand{\tensakuClean}{
    \renewcommand{\TSK}[1]{}
}
\newcommand{\TSK}[1]{#1}

\newcommand{\mkclean}{
    \renewcommand{\reminder}[1]{}
    \renewcommand{\duty}[1]{}
}

\newcommand{\vswd}{\vspace{0.3em}}
%\newcommand{\vswd}{\vspace{0.0em}}
%%%%%%%%%%%%%%%%%%%%%%%%%%%%%%%%%%%%%%%%%
% from cf: shorthands - also they make tighter lists
%\newcommand{\bit}{\vspace{-0.5em}\begin{itemize*}}
\newcommand{\bit}{\vswd\begin{itemize*}}
\newcommand{\eit}{\end{itemize*}\vswd}
\newcommand{\ben}{\vswd\begin{enumerate*}}
\newcommand{\een}{\end{enumerate*}\vswd}
\newcommand{\bea}{\vspace{-0.0em}\begin{eqnarray}}
\newcommand{\eea}{\end{eqnarray}\vspace{-0.0em}}
\newcommand{\beq}{\vspace{-0.0em}\begin{equation}}
\newcommand{\eeq}{\end{equation}\vspace{-0.0em}}

\renewcommand{\bit}{\vswd\begin{compactitem}}
\renewcommand{\eit}{\end{compactitem}\vswd}
\renewcommand{\ben}{\vswd\begin{compactenum}}
\renewcommand{\een}{\end{compactenum}\vswd}

\newcommand{\argmax}{\mathop{\rm arg~max}\limits}
\newcommand{\argmin}{\mathop{\rm arg~min}\limits}

\newcommand{\mrk}{$\surd$}  % check mark

\newcommand{\emp}{\bf \underline}
\newcommand{\bi}{\bfseries\itshape} 
\newcommand{\ac}[1]{\acute{#1}}

% for the properties of our model G1, G2, G3, ...
\newcommand{\prop}[1]{({\bf{#1}})\xspace}
\newcommand{\propn}[1]{({{#1}})\xspace}

\newcommand{\googletn}{{GoogleTrend}\xspace}
\newcommand{\mocapn}{{MoCap}\xspace}
\newcommand{\googlet}{{\it \googletn}\xspace}
\newcommand{\mocap}{{\it \mocapn}\xspace}

% RegimeCast, DataVane
\newcommand{\methodn}{\it SplitSEE}
\newcommand{\methodN}{\-Split\-SEE\xspace}
\newcommand{\method}{\textsc{\it SplitSEE}\xspace}
\newcommand{\methodSSn}{Epi\-Snap\xspace}
\newcommand{\methodSS}{\textsc{\methodSSn}\xspace}
\newcommand{\modelsearchn}{{Epi\-Finder\xspace}}
\newcommand{\modelestimatorn}{{Epi\-Estimator\xspace}}
\newcommand{\modelsearch}{\textsc{\modelsearchn}\xspace}
\newcommand{\modelestimator}{\textsc{\modelestimatorn}\xspace}
\newcommand{\regime}{regime\xspace}
\newcommand{\Regime}{Regime\xspace}
\newcommand{\regimes}{regimes\xspace}
\newcommand{\regimeshift}{regime shift\xspace}
\newcommand{\regimeshifts}{regime shifts\xspace}
\newcommand{\Regimeshift}{Regime shift\xspace}
\newcommand{\Regimeshifts}{Regime shifts\xspace}

\newcommand{\currentwindow}{current window\xspace}
\newcommand{\Currentwindow}{Current window\xspace}
\newcommand{\forecastwindow}{forecast window\xspace}
\newcommand{\Forecastwindow}{Forecast window\xspace}
\newcommand{\eventstream}{epidemic stream\xspace}
\newcommand{\eventstreams}{epidemic streams\xspace}

\newcommand{\vct}{\boldsymbol{ct}}
\newcommand{\vx}{\boldsymbol{x}}
\newcommand{\vz}{\boldsymbol{z}}
\newcommand{\vc}{\boldsymbol{c}}
\newcommand{\vp}{\boldsymbol{p}}
\newcommand{\vq}{\boldsymbol{q}}
\newcommand{\ve}{\boldsymbol{e}}
\newcommand{\vX}{\boldsymbol{X}}
\newcommand{\vZ}{\boldsymbol{Z}}
\newcommand{\vC}{\boldsymbol{C}}
\newcommand{\vQ}{\boldsymbol{Q}}
\newcommand{\prob}[2]{P(#1|#2)}
\newcommand{\X}{\mathcal{X}}
\newcommand{\Y}{\mathcal{Y}}
\newcommand{\sumk}[1]{\sum_{k=#1}^K}
\newcommand{\mynorm}[2]{\left|\!\left|#1\right|\!\right|_{#2}}
\newcommand*\samethanks[1][\value{footnote}]{\footnotemark[#1]}
\newcommand{\myparaitemize}[1]{\noindent{\textbf{#1.}}}

% Country names
\newcommand{\czechia}{Czechia\xspace}

\newcommand{\macromap}{\method-Map\xspace}

%
%competitors
\newcommand{\plif}{\textsc{PLiF}\xspace}
\newcommand{\LV}{\textsc{LVC}\xspace}
\newcommand{\ecoweb}{\textsc{EcoWeb}\xspace}
\newcommand{\funnel}{\textsc{FUNNEL}\xspace}
\newcommand{\AR}{\textsc{AR}\xspace}
\newcommand{\var}{\textsc{VAR}\xspace}
\newcommand{\arima}{\textsc{ARIMA}\xspace}
\newcommand{\tbats}{\textsc{TBATS}\xspace}
\newcommand{\sird}{\textsc{SIRD}\xspace}
\newcommand{\gru}{\textsc{GRU}\xspace}
\newcommand{\deepar}{\textsc{DeepAR}\xspace}
\newcommand{\epideep}{\textsc{EpiDeep}\xspace}
\newcommand{\tcn}{\textsc{TCN}\xspace}
\newcommand{\tstcc}{\textsc{TS-TCC}\xspace}
%

% --- tensor  --- %
%\DeclareMathAlphabet{\mathpzc}{OT1}{pzc}{m}{it}
%\DeclareFontFamily{OT1}{pzc}{}
%\DeclareFontShape{OT1}{pzc}{m}{it}{<-> s * [1.200] pzcmi7t}{}
%\DeclareMathAlphabet{\mathpzc}{OT1}{pzc}{m}{it}
%\newcommand{\tensor}[1]{\mathpzc{#1}}
%
\newcommand{\mathN}{\mathbb{N}}
\newcommand{\tenX}{\mathcal{X}}
\newcommand{\Xf}{{\mathcal{X}_F}}   % data 
\newcommand{\Xc}{{\mathcal{X}_C}}   % data 
\newcommand{\Xtmp}{X_C}
\newcommand{\Xtmpi}{X_C^{(i)}}
\newcommand{\xb}{\bm{x}}   % data
\newcommand{\x}{x}   % data
\newcommand{\solution}{\mathcal{M}}   % models

\newcommand{\Vs}{\mathcal{V}}
\newcommand{\Vstmp}{{V}}
\newcommand{\Vstmpi}{{V^{(i)}}}
\newcommand{\vb}{\mathbf{v}}
\newcommand{\Vc}{\mathcal{V}_C}
\newcommand{\Vf}{{V}_{\tmed+\nstep}}
\newcommand{\Ve}{\mathcal{V}_E}
\newcommand{\Vetmp}{{V}_E}   %
\newcommand{\Vetmpi}{{V}_E^{(i)}}   %
\newcommand{\Vctmp}{{V_C}}
\newcommand{\Vctmpi}{{V_C^{(i)}}}
\newcommand{\Ws}{{W}}
\newcommand{\We}{{W_E}}   % 
\newcommand{\Ss}{{S}}
\newcommand{\Se}{{S_E}}   % 
\newcommand{\Ses}{\hat{S}_{E}}   %
\newcommand{\dps}{\delta}   %
\newcommand{\StaNM}{Potential activity\xspace}
\newcommand{\staNM}{potential activity\xspace}
\newcommand{\ObsNM}{Estimated event\xspace}   
\newcommand{\obsNM}{estimated event\xspace}   
\newcommand{\WNM}{\Regime activity\xspace} 
\newcommand{\wNM}{\regime activity\xspace} 

\newcommand{\Obs}{\bm{v}}   % observations
\newcommand{\Sta}{\bm{s}}   % states
\newcommand{\StaS}{\mathbf{S}}   % states^2
\newcommand{\obs}{{v}}   % obs
\newcommand{\sta}{{s}}   % sta 
\newcommand{\MQs}{\bm{\Theta}}   % model params 
\newcommand{\MQsE}{\bm{\Theta}^E}   % model params 
\newcommand{\MQsL}{\bm{\Theta}^L}   % model params 
\newcommand{\mparam}{\bm{\theta}}   % models
\newcommand{\lparam}{\bm{\theta^{L}}}
\newcommand{\eparam}{\bm{\theta^{E}}}
\newcommand{\mparamc}{\bm{\theta}_C}   % models
\newcommand{\lsetn}{location parameters\xspace}
\newcommand{\esetn}{epidemic parameters\xspace}
\newcommand{\nmodels}{g}
\newcommand{\Candi}{\mathcal{C}}   % model params 
\newcommand{\rM}{\mathbf{R}} 
\newcommand{\rT}{\mathcal{R}} 
%\newcommand{\wM}{\mathbf{W}}   % weight 

% SEIRD model
\newcommand{\susceptible}{S}
\newcommand{\exposed}{E}
\newcommand{\infectious}{I}
\newcommand{\recovered}{R}
\newcommand{\dead}{D}
\newcommand{\exposedi}{E_0}
\newcommand{\infectiousi}{I_0}
\newcommand{\recoveredi}{R_0}
\newcommand{\deadi}{D_0}
\newcommand{\stpoint}{t_0}  % start point

\newcommand{\infect}{\beta}
\newcommand{\infecti}{\beta_{0}}
\newcommand{\exponential}{\sigma}
\newcommand{\exponentiali}{\sigma{0}}
\newcommand{\heal}{\delta}
\newcommand{\forget}{\gamma}
\newcommand{\vaccine}{\theta}
\newcommand{\vaccinei}{\theta_{0}}
\newcommand{\death}{\delta}
\newcommand{\Nmax}{N} % # of potential bloggers
\newcommand{\betaN}{\beta * \Nmax}
\newcommand{\tv}{t_{\vaccine}}
\newcommand{\tc}{t_c}
\newcommand{\Sc}{S_c}
\newcommand{\bgn}{\epsilon}
\newcommand{\pfreq}{P_p}
\newcommand{\prate}{P_a} %%{P_{low}} 
\newcommand{\pshift}{P_s}

\newcommand{\wb}{\bm{w}}   % weight
\newcommand{\w}{{w}}   % weight
\newcommand{\V}{V}   
\newcommand{\func}{f}   
\newcommand{\ma}{g}   
\newcommand{\Sloc}{S}

\newcommand{\rms}{f_{RMS}}
\newcommand{\Stai}{\bm{s_0}} 
\newcommand{\aV}{\bm{p}} 
\newcommand{\aM}{\mathbf{Q}}
\newcommand{\aT}{{\mathcal A}}
\newcommand{\aTe}{{a}}
\newcommand{\bV}{\bm{u}} 
\newcommand{\bM}{\mathbf{V}}
\newcommand{\rV}{\bm{r}} 
\newcommand{\Hr}[2]{{#1}^{(#2)}}
\newcommand{\Hi}[1]{^{(#1)}}

\newcommand{\geoschem}{local area\xspace}
\newcommand{\geoschems}{local areas\xspace}
\newcommand{\Geoschems}{Local areas\xspace}

\newcommand{\nclass}{k}
\newcommand{\height}{h} %|H|}
\newcommand{\Hs}{H}
\newcommand{\nregimes}{c}
% window
\newcommand{\ttick}{t}
\newcommand{\tmst}{t_{m}}
\newcommand{\tmed}{t_{c}}
\newcommand{\tfst}{t_{s}}
\newcommand{\tfed}{t_{e}}
\newcommand{\lene}{l_{e}} %l_{c}}
\newcommand{\lenc}{l_{c}}
\newcommand{\lena}{l_{s}}
\newcommand{\lens}{l_{p}}

\newcommand{\duration}{\tmed}
\newcommand{\ndims}{d}
\newcommand{\nlocs}{r}
\newcommand{\ngeoschems}{h}

\newcommand{\nstep}{\lena\xspace}
\newcommand{\slwd}{\lens\xspace}
\newcommand{\nstepA}{$\nstep$-steps-ahead\xspace}

\newtheorem{expectation}{EXPECTATION}
\newtheorem{informalProblem}{Informal Problem}
\newtheorem{lemma}{Lemma}
\newtheorem{theorem}{Theorem}
\newtheorem{problem}{Problem}
\newtheorem{mathmodel}{Model}

\newcommand{\covid}{COVID-19\xspace}
\renewcommand{\algorithmicrequire}{\textbf{Input:}}
\renewcommand{\algorithmicensure}{\textbf{Output:}}

\newcommand{\adabracket}[1]{\left(#1\right)}
\newcommand{\adarecbracket}[1]{\left[#1\right]}
\newcommand{\entropy}[1]{\widehat{\mathcal{H}}\adabracket{#1}}
\newcommand{\approxEntropy}[1]{\widehat{\mathcal{H}}\adabracket{#1}}

\newcommand{\vy}{\boldsymbol{y}}
% \newcommand{\vx}{\boldsymbol{x}}
% \newcommand{\vz}{\boldsymbol{z}}
% \newcommand{\vc}{\boldsymbol{c}}
% \newcommand{\vp}{\boldsymbol{p}}
% \newcommand{\vq}{\boldsymbol{q}}
% \newcommand{\ve}{\boldsymbol{e}}
% \newcommand{\vX}{\boldsymbol{X}}
% \newcommand{\vZ}{\boldsymbol{Z}}
% \newcommand{\vC}{\boldsymbol{C}}
% \newcommand{\vQ}{\boldsymbol{Q}}
% \newcommand{\prob}[2]{P(#1|#2)}
% % \newcommand{\X}{\mathcal{X}}
% \newcommand{\Y}{\mathcal{Y}}
% \newcommand{\sumk}[1]{\sum_{k=#1}^K}
% \newcommand{\mynorm}[2]{\left|\!\left|#1\right|\!\right|_{#2}}
% \newcommand*\samethanks[1][\value{footnote}]{\footnotemark[#1]}
\begin{abstract}
  While end-to-end multi-channel electroencephalography (EEG) learning approaches have shown significant promise, their applicability is often constrained in  neurological diagnostics, such as intracranial EEG resources.
When provided with a single-channel EEG, how can we learn representations that are robust to multi-channels and scalable across varied tasks, such as seizure prediction?
In this paper, we present \method, a structurally splittable framework designed for effective temporal-frequency representation learning in single-channel EEG.
The key concept of \method is a self-supervised framework incorporating a deep clustering task. 
Given an EEG, we argue that the time and frequency domains are two distinct perspectives, and hence, learned representations should share the same cluster assignment.
To this end, we first propose two domain-specific modules that independently learn domain-specific representation and address the temporal-frequency tradeoff issue in conventional spectrogram-based methods. 
Then, we introduce a novel clustering loss to measure the information similarity.
This encourages representations from both domains to coherently describe the same input by assigning them a consistent cluster. 
\method leverages a pre-training-to-fine-tuning framework within a splittable architecture and has following properties:
(a) {\it Effectiveness}:
it learns representations solely from single-channel EEG but has even outperformed multi-channel baselines.
(b) {\it Robustness}: 
it shows the capacity to adapt across different channels with low performance variance. Superior performance is also achieved with our collected clinical dataset.
(c) {\it Scalability}:
With just \emph{one fine-tuning epoch}, \method achieves high and stable performance using partial model layers.
\end{abstract}

\section{Introduction}
    \label{sec:intro}
    
\IEEEPARstart{E}{lectroencephalography} (EEG) reveals the activities of millions of neurons in multi-channel recordings.
As a staple brain imaging tool, it is now widely used in various healthcare domains, including sleep medicine \cite{TSTCC_ijcai21, chenIJCAI,SleepTransformer} and seizure epilepsy detection \cite{GNN_ICLR22}.
% and rehabilitation systems \cite{2011_TMBE_ARM, PNASRL,rehabilitation}.
Deep learning models have shown impressive success in automating EEG analysis.
Since meaningful features have region-specific activation effects, successful methods typically involve a \emph{multi-channel modeling framework} \cite{MBrain_KDD23, Chen_SDM23, GNN_AAAI23}.
% Successful methods typically involve a \emph{multi-channel modeling framework} \cite{MBrain_KDD23, Chen_SDM23, GNN_AAAI23}.
Their main objective is to learn holistic topographic brain information for localizing eventful features \cite{DomainAdaptaion_AAAI2021}.
Learning such spatial or spatio-temporal information has shown great potential in EEG analyses \cite{Rasheed,IAG_AAAI2021, pmlrEEG}. 
Despite their success, there is a need in neurological diagnostics \cite{KDDEEG1, AttnSleep_21}, particularly in disease detection and portable monitoring, for developing accurate and disruption-reduced methods using fewer or even \emph{single channels}, specifically,

\begin{itemize}[left=0pt]
    \item Multi-channel methods are often inapplicable to some data resources, such as intracranial EEG, which is limited to specific brain regions, lacking holistic multi-channel recordings, but important for pre-surgical analysis \cite{Kishima_ieeg}.
    \item Continuous wear is essential for diagnostics but is impractical with multi-channel EEG because it often disrupts natural conditions, such as sleep \cite{SeqSleepNet}.
\end{itemize}

\begin{figure}[t]
    \centering
    \includegraphics[width=0.96\linewidth]{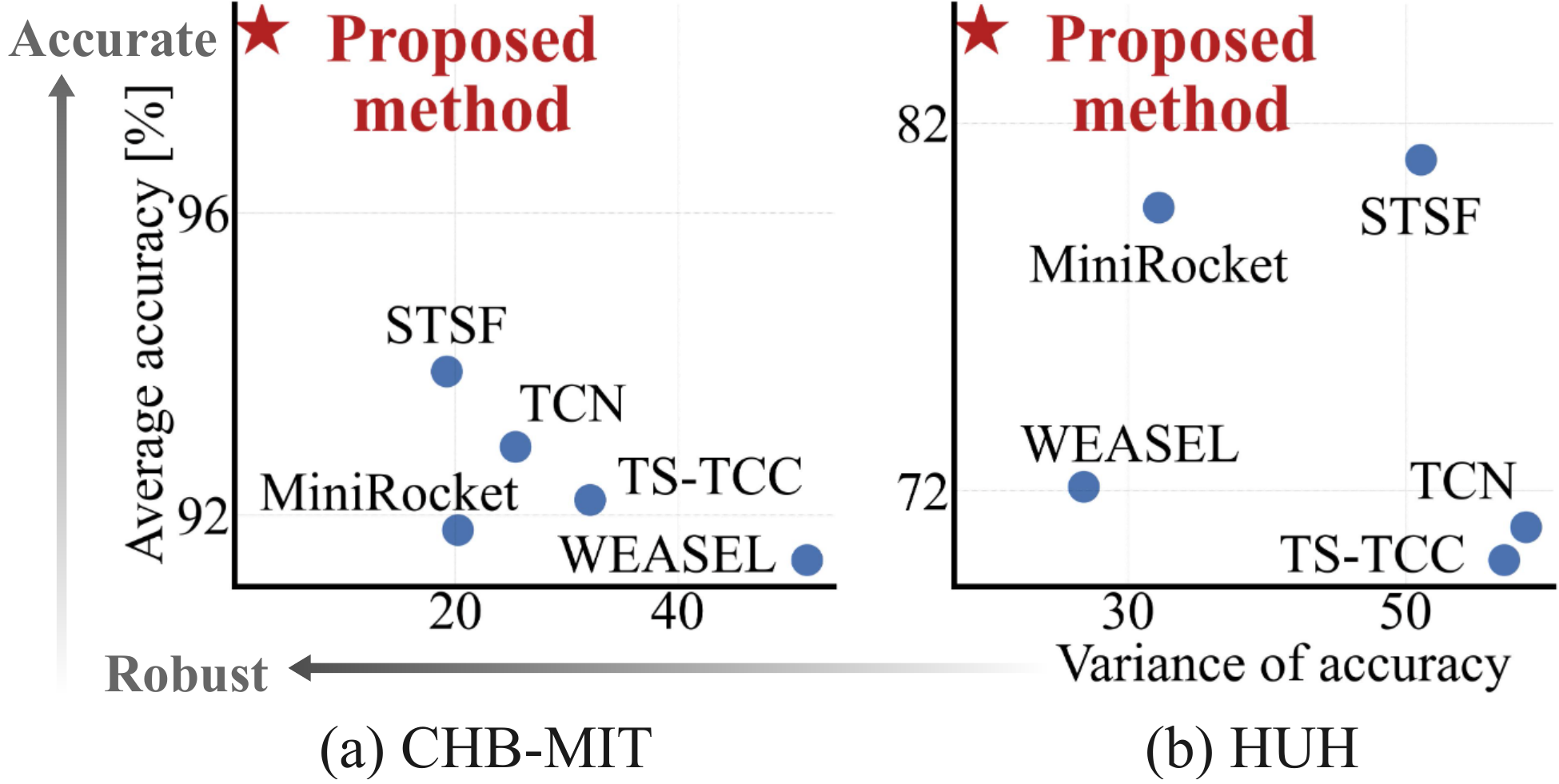} %\\
    \vspace{-0.4em} 
    \caption{The scatter plot presents the seizure prediction results of \method compared with single-channel baselines on two public EEG datasets. 
    Points located in the upper-left quadrant indicate that our single-channel method maintains robust performance across various EEG inputs with lower variance and higher average accuracy. 
    These results underscore the superior performance and reliability of \method in seizure prediction.}
    \vspace{-0.4em}
    \label{fig:scatter}
    \vspace{-0.4em}
\end{figure}

In this paper, we aim to devise \emph{a flexible training strategy that can effectively capture meaningful features solely from \textbf{single-channel EEG} for varying channels and tasks.}
Recently, several single-channel EEG methods have been proposed \cite{Qu_nn1, SleepKD_ijicai23, ChenTNSRE}.
Since lacking spatial information, these methods aim to learn temporal-frequency features for specific tasks;
however, some limitations remain.

\noindent \textbf{1) Inadequate feature encoding and trade-off issue in temporal-frequency learning.}
    Existing works can generally be categorized into two approaches: using deep learning models on raw EEG data \cite{TSTCC_ijcai21,MBrain_KDD23} and preprocessing the data with short-time Fourier transform (STFT) and learning features on the resulting spectrogram \cite{chenIJCAI,SleepTransformer}.
    The former methods involve feature fusion from multiple encoders for different temporal resolutions.
    However, meaningful features, manifested as various frequency waves, often are intermixed in EEG signals \cite{NEURIPS2019EEG}.
    The learned features are often frequency-entangled.
    Moreover, researchers use the spectrogram to capture temporal-frequency features and their correlations.
    The trade-off between temporal and frequency resolution is a long-standing problem in spectrogram learning \cite{time-frequency, Deconvolutive_STFT}.

\noindent \textbf{2) Task-specific end-to-end training.} Existing methods rely heavily on a supervised learning framework, necessitating vast amounts of high-quality labeled data for end-to-end model training \cite{ACMTransCH}.
    Facing newly collected data, a de-novo model training is needed.
    These methods are channel or task-specific, often with low generalization to other tasks \cite{Chen_SDM23}.
    Therefore, is it possible to develop a model that is applicable to both seizure detection and sleep staging?
    Can knowledge be transferred from an EEG dataset to another hospital dataset?
    In reality, procuring labels is time-consuming \cite{KDDEEG1}.
    It is crucial to explore the potential of utilizing various unlabeled data sources.

Therefore, we propose \textbf{\emph{SplitSEE}}, a $\mathbf{Split}$table learning framework tailored for $\mathbf{S}$ingle-channel $\mathbf{EE}$G representation with the following contributions:

\noindent \textbf{1) Deep clustering for independent temporal-frequency representations.}
    Considering the time and frequency domains as two distinct perspectives of EEG data, in this paper, we formulate temporal-frequency learning as a clustering task.
    Specifically, given an input, representations learned from both the time and frequency domains should share the same cluster assignment. 
    Hence, \method involves (i) two domain-specific feature learning modules and (ii) cross-domain alignment, i.e., clustering. 
    The domain-specific modules independently learn features of both time and frequency with multi-granularities.
    For cross-domain alignment, we introduce a novel loss function inspired by recent deep clustering methods \cite{SwAV}.
    This loss encourages a consistent cluster assignment for temporal and frequency features, ensuring that these two features coherently describe the same input data.
    These designs mitigate the trade-off issue and enhance representation quality for eventful features. 
    To our knowledge, \method is the first to formulate temporal-frequency learning as a clustering task.

\noindent \textbf{2) Domain-specific self-supervised learning framework with a splittable architecture.} 
    The driving force behind \method is self-supervised learning (SSL).
    SSL has recently drawn attention to data representation by employing rich unlabeled data \cite{TSTCC_ijcai21,pmlrEEG, Chen_SDM23, MBrain_KDD23}.
    Firstly, our temporal-frequency clustering is designed using SSL.
    Secondly, we propose two module-specific contrastive learning (CL), which facilitate two independent representation learning modules.
    Each module is well-pre-trained by CL, and architecturally, the structure of \method is splittable: only selected layers are used for fine-tuning with label supervision and serving downstream tasks.
    This eliminates the need for end-to-end model re-training.

% \mypara{Preview of results}
We evaluate \method on five datasets: four public datasets and \textbf{our self-collected clinical dataset}, across two EEG tasks.
Our method has the following properties:
\underline{Effectiveness}:
\method learns representations solely from single-channel EEG but has even outperformed multi-channel baselines.
\underline{Robustness}: 
\method has the capacity to adapt different channels with low performance variance. 
Fig. \ref{fig:scatter} provides a preview result.
\underline{Scalability}:
Our experiments show that with just \emph{one fine-tuning epoch}, \method achieves high and stable performance using partial model layers.
We further develop a \ul{split federated learning version} of \method, wherein pre-training is handled on the server, with fine-tuning executed locally.
\section{Related work}
    \label{sec:related}
    
\subsection{Single-Channel EEG Studies}\label{subsec:singlelearning}

Deploying too many channels can potentially increase the risk of skin irritation and discomfort \cite{MS-HNN_23}. 
Some studies are considering the use of fewer channels to minimize noise and computational expenses \cite{Qu_nn1, GNN_ICLR22, GNN_AAAI23}.
To be more effective, some studies \cite{DeepSleepNet, DeepSleepNet-Lite, TinySleepNet} consider modeling a single-channel EEG; for example, AttnSleep \cite{AttnSleep_21} deploys causal convolutions to model the temporal relations of the single-channel input for sleep staging.
MS-HNN \cite{MS-HNN_23} proposes a multi-scale learning framework that uses CNN and RNN to extract more beneficial features from single-channel EEGs across various frequencies. 
A seizure detection work \cite{ChenTNSRE} refines the temporal-spatial features and models the correlations between them with the Transformer model.

\subsection{Self-Supervised Learning in EEGs}\label{subsec:EEG}

Early attempts in SSL adopted on the unsupervised pre-trained sequential model, such as BERT \cite{BERT_NAACL19}, to model the long-term temporal dependencies of EEG data for fine-tuning \cite{TARNet_kdd22}.
Most methods focus primarily on modeling multi-channel EEGs \cite{GNN_ICLR22,MBrain_KDD23}.
For single-channel EEG, TS-TCC \cite{TSTCC_ijcai21}, a contrastive learning (CL) framework (one of method in SSL) for time-series representation, introduces novel data augmentation techniques and cross-view prediction tasks tailored for time-series data, including EEGs.
However, this method does not incorporate frequency domain features for EEG representation. 
In response, BTSF \cite{BTSF} and TF-C \cite{TF-C} have addressed this gap and emphasize temporal-frequency fusion. 
Recently, \cite{Chen_SDM23} has explored a temporal-frequency method and reported superior results in two EEG tasks. 
However, the SOTA works \cite{ICLRcost,GNN_AAAI23,Chen_SDM23} are typically multi-channel SSL frameworks; moreover, \method learns meaningful temporal-frequency representation by formulating a deep clustering in an SSL fashion.

\section{Problem Setting}
    \label{sec:preliminary}
    \label{section:pre}

\subsection{Single-Channel EEG Data}
EEG data consist of multi-channel recordings, where each channel captures a series of neuronal activities from a specific brain region over time.
Formally, let $\mathbf{X} = {\{\mathbf{X}^{(m)}\}}_{m=1}^{C}$ denote EEG data composed of $C$ channels.
Here, $\mathbf{X}^{(m)} \in \mathbb{R}^{N \times L}$ indicates a single-channel data that collects $N$ subjects over a duration of $L$ time points.
For clarity, each $\mathbf{X}^{(m)} \in \mathbb{R}^{N \times L}$ can be expressed as a sequence of vectors: $(\vx_1, \vx_2, \dots, \vx_N)$.
Let $\mathbf{Y} = \{\vy_1, \vy_2, \cdots, \vy_{|\mathbf{Y}|}\}$ denotes the set of labels associated with $\mathbf{X}^{(m)}$. For instance, $\mathbf{Y}$ could represent either the types of sleep stages or a binary normal/seizure classification.
The symbols and definitions are described in TABLE \ref{notation}.

\subsection{Problem Formulation}
\label{subsec:problem}

Our primary objective is to build a well-performing model $f_{\theta}(\cdot)$, which functions with a single-channel EEG $\vx_i$, where $i = 1, 2, \dots , N$. 
The aim is to learn a lower dimensional vector $\vz_i$ that represents all essential information across all $C$ channels for $\mathbf{X_i} = (\vx_i^{(1)}, \vx_i^{(2)}, \dots , \vx_i^{(C)})$.
That is, using $\vz_i$ and $\vx_i$ is equivalent for any downstream task. For even greater effectiveness, $\vz_i$ should ideally exhibit performance that is competitive with the multi-channel learning methods in $\{\mathbf{X}_i\}_{m=1}^{C}$.
Practically, $\vz$ is produced by $\vz = f_{\theta}(\vx)$, where the function is parameterized by $\theta$.
To obtain reliable $\vz$, we have three expectations for $f_{\theta}(\cdot)$:

\begin{expectation}
Operate on single-channel EEG data, with the capacity to perform competitively with multi-channel or multi-data source modeling.
\end{expectation}

\begin{expectation}
Learn high-quality representations $\vz$ from $\vx$ without label supervision, which is applicable to various EEG tasks across different channels. 
\end{expectation}

\begin{expectation}
Develop a flexible framework to overcome the constraints of end-to-end model re-training and enhance its practical utility in real-world clinical settings with limited computing resources.
\end{expectation}

\noindent We confirm the realization of the above expectations in Sections \ref{sec:result}.

\begin{table}[t]
    \caption{
    Symbols and definitions.}
    \centering
    \label{notation}
    \resizebox{0.95\linewidth}{!}{%
        \begin{tabular}{l|l}
            \toprule
            \textbf{Symbol}    & \textbf{Definition} \\
            \midrule
            $\mathbf{X}$       & EEG data, i.e., $\mathbf{X} \in \mathbb{R}^{C \times N \times L}$  \\
            $C$ & EEG channel dimensions      \\
            $N$ & Number of subjects      \\
            $L$ & Number of time steps       \\
            $\mathbf{Y}$ & Label set of downstream task \\
            \midrule
            $\vx^{\scriptscriptstyle \text{weak}}, \vx^{\scriptscriptstyle \text{strong}}, \vx^{\scriptscriptstyle \text{abs}}$ &  Observation view\\%, $\vx_{\text{weak}} \in \mathbb{R}^{L}$ \\
            $\vZ^w, \vZ^s, \vZ^l, \vZ^h$& Local features from each view\\%, $\mathbf{Z}^w \in \mathbb{R}^{L \times D}$ 
            $\vct^w, \vct^s, \vct^l, \vct^h$ & Global features from each view \\
            \midrule
            $f_{\theta}(\cdot)$ & Proposed \method function       \\
            $TCN(\cdot)$ & TCN function       \\
            $CNN_{1d}(\cdot)$ & 1-dimensional CNN function       \\
            $Trans_{T}(\cdot)$, $Trans_{F}(\cdot)$ & Transformer encoder function of each domain      \\
            $L^s_{TC}$, $L^w_{TC}$, $L^l_{FC}$, $L^h_{FC}$  & Loss functions of each view \\
            $L_{dc}$ & Loss functions of temporal-frequency clustering \\
            \bottomrule
        \end{tabular}
    }
\end{table}
\section{Proposed method}
    \label{sec:moethod}
    \begin{figure*}[t]
    \centering
\includegraphics[width=0.98\linewidth]{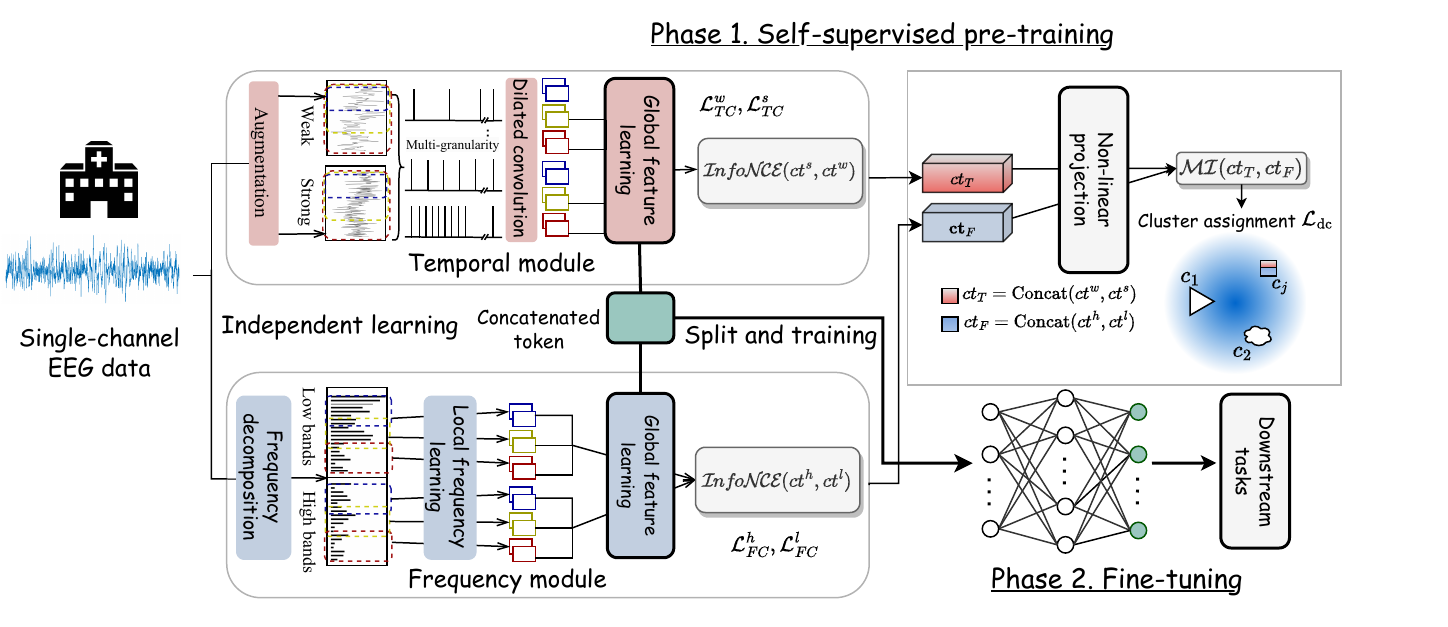} %\\
	    \vspace{-0.20em} \\
\vspace{-0.20em}
    \caption{
    \textbf{\emph{SplitSEE}} overview. The architecture integrates a two-phase learning approach: an initial self-supervised pre-training phase followed by a fine-tuning phase. It comprises three representation learning modules: two domain-specific feature learning modules for temporal-frequency independent learning, and a cross-domain feature clustering module. 
    Moreover, the splittable architecture allows seamless adaptation to various tasks using selected layers. 
    Each domain-specific module is trained by InfoNCE loss function, and the deep clustering for alignment is performed by maximizing mutual information (MI).}
    
    \label{fig:overview}
\vspace{-0.20em}
\end{figure*}

\subsection{Forward Process Overview}
\label{subsec:forward}

We develop a deep clustering framework to train $f_{\theta}(\cdot)$, employing a two-step process to obtain a representative vector $\vz$ for a single-channel EEG $\vx$. 
Fig. \ref{fig:overview} shows workflow of \method.
Fundamentally, \method consists of three key modules: two dedicated to learning in the temporal and frequency domains, and a deep clustering module designed to align the learned features. 
Initially, $\vx$ is processed independently by the domain-specific modules to extract temporal ($\vct_T$) and frequency ($\vct_F$) representations (see Sections \ref{subsec:TFindependentlearning} and \ref{subsec:frequencymodule}). These representations are then input into the deep clustering module. This module optimizes the representations to ensure they share the same cluster assignment, which integrates them into a unified feature space (see Section \ref{subsec:alignmodule}).

\noindent\textbf{Remark.}
Two independent domain-specific modules aim to address the limitations of conventional methods, such as the trade-offs associated with spectrogram usage and inadequate temporal-frequency encoding.
Each module is designed in a CL manner.
This ensures the learned features are disentangled and more relevant to their respective domains.
Further clustering is an information metric, making the temporal-frequency representations coherently describe the same input data.

\subsection{Multi-Granularity Temporal Independent Learning}\label{subsec:TFindependentlearning}

\emph{Motivation.}
In general, EEG data lack recognizable patterns, e.g., trend and seasonality.  
The eventful features are transient and temporally unpredictable. 
For instance, a spontaneous K-complex waveform often exhibits bursts within 0.5$\sim$1.5 seconds \cite{ChenTNSRE}.
Therefore, this module involves multi-granularity learning that encodes different levels of temporal dependencies.

\subsubsection{Data augmentation}
We initially generate two noisy inputs of the single-channel EEG data.
Given $\vx$, we introduce random noises to the signal and adjust its amplitude for weak augmentation. 
For strong augmentation, we divided the input EEG data into a random set of segments, capped at a total of $L$.
We randomly shuffle these segments.
The objective is to increase robustness for \emph{capturing millisecond-level key features that appear randomly.}
We limit the number of cut points to a maximum of 4 to 11, ensuring that most of EEG waveform within the entire time points is preserved.
These augmentations are formulated as:
\begin{align}
\vx^{\scriptscriptstyle \text{weak}} = \alpha (\vx + \epsilon_w), 
\quad  \vx^{\scriptscriptstyle \text{strong}} = \{\vx_{\beta(1)}, \vx_{\beta(2)}, ..., \vx_{\beta(L)}\}+ \epsilon_s,
\end{align}
where $\epsilon_w$ and $\epsilon_s$ denote the respective random noises, $\alpha$ is a scaling factor and $\beta(\cdot)$ is a random permutation function.

\subsubsection{Multi-granularity feature extraction}
The backbone is a temporal convolutional network (TCN) \cite{TCN} with dilated convolutional layers.
By introducing a fixed step between successive filter taps, we expand the receptive field of different EEG scales of EEG data points as:
\begin{align}
    \vZ^w, \vZ^s = \text{TCN}(\vx^{\scriptscriptstyle \text{weak}}, \vx^{\scriptscriptstyle \text{strong}})
\end{align}
where each $\vZ^w, \vZ^s \in \mathbb{R}^{L \times |z|}$  is characterized as $[\vz_{1}, \vz_{2}, \dots \vz_{L}]$, 
$L$ is the number of timesteps. $|z|$ represents feature embedding dimensions in different granularity.

Unlike previous works \cite{TSTCC_ijcai21,Chen_SDM23} that fuse and preserve filtered features as a vector, our method expands ($\vx \rightarrow \vZ$) by preserving a landscape of local features at various granularities, such as 0.5-second and 1-second intervals, each characterized by distinct receptive fields.

\subsubsection{Temporal semantic information summarization}
We employ the Transformer encoder \cite{ChenTNSRE} with an additional class token to extrapolate the global long-term dependencies in EEGs. 
We first segment the $\mathbf{Z}^{w}$ and $\mathbf{Z}^{s}$ sequence into $T$ patches, that is, $\mathbf{Z}^{w} = (\vz_{1}^{w} , \vz_{2}^{w} , \dots z_{T}^{w})$ and $\mathbf{Z}^{s} = (\vz_{1}^{s}, \vz_{2}^{s}, \dots \vz_{T}^{s})$.
Then we fed them into the Transformer encoders $\text{Trans}_T$, and summarized context class tokens, denoted as: 

\begin{align}
& \vct^{w}, \vct^{s} = \text{Trans}_T([\vz_{1}^{w}, \vz_{2}^{w}, \dots \vz_{T}^{w}], [\vz_{1}^{s}, \vz_{2}^{s}, \dots \vz_{T}^{s}])
\end{align}

\subsubsection{Temporal Contrastive Learning} 
We conduct dual contrastive learning to capture the temporal dynamics in EEG.
This establishes the past-future prediction, aiming to tackle the randomness of informative EEG waveforms.
The dual or bi-prediction increases the observations for the randomness.

Formally, we use $\vct^{w}$ to predict $\vz^{s}_{T+K}$ $(1 < K \leq L-T)$ and $\vct^{s}$ to predict $\vz^{w}_{T+K}$, where $K$ is randomly chosen. 
We employ linear projection layers $W^K_T(\cdot)$ following the Transformer encoder as our autoregressive model.
The similar pairs, \textit{positive samples}, are $\vz_{T+K}^{s}$ and $\vz_{T+K}^{w}$ for $W^K_T(\vct^{w})$ and $W^K_T(\vct^{s})$.
Conversely, the dissimilar pairs $\vz_n$, known as \textit{negative samples}, consist of samples from the mini-batch.
The contrastive losses $L^s_{TC}$ and $L_{TC}^{w}$ are calculated as the Information Noise Contrastive Estimation (InfoNCE) loss \cite{Belghazi2018-MINE} as follows:

\begin{align}
    L^s_{TC}:= \mathbb{E}_{\mathcal{X}}\adarecbracket{-\log \frac{\exp\adabracket{(W^K_T(\vct^{s})) \cdot \vz_{T+K}^w) / \tau}}{\sum_{n}\exp\adabracket{(W^K_T(\vct^{s}))\cdot \vz_{n}^w / \tau} }},\\
   L^w_{TC}:= \mathbb{E}_{\mathcal{X}}\adarecbracket{-\log \frac{\exp\adabracket{(W^K_T(\vct^{w})) \cdot \vz_{T+K}^s / \tau}}{\sum_{n}\exp\adabracket{(W^K_T(\vct^{w})) \cdot \vz_{n}^s / \tau} }}
\end{align}

\subsection{Multi-Granularity Frequency Independent Learning}\label{subsec:frequencymodule}
\emph{Motivation.}
To learn high-quality frequency representation, we propose this module that independently learns eventful features from the full frequency spectrum with precise resolution.

\subsubsection{Frequency decomposition}
We apply the Fast Fourier Transform (FFT) to the normalized signal $\vx$ to decompose it into its frequency spectrum. 
We then extract the absolute values from the mirrored first half of the frequency bands, denoted as $\vx^{\text{FFT}} = |\text{FFT}(\text{Norm}(\vx))|$.

\subsubsection{Frequency local feature extraction}
We use a 1D convolution block to learn the intricate bi-directional correlations between narrow frequency bands, given by:
\begin{align}
\vZ^l= \text{CNN}_{1d}(\vx^{\scriptstyle \text{FFT}}), \vZ^l \in \mathbb{R}^{B \times |z|}.
\end{align}
where $B$ is the CNN down-sampled dimensions of frequency bands. 
We create the feature vector $\vZ^h$ by reversing the order of elements in the vector $\vZ^l$.
\begin{align}
\vct^{h}, \vct^{l} = \text{Trans}_F([\vz_{1}^{l}, \vz_{2}^{l}, \dots \vz_{F}^{l}], [\vz_{1}^{h}, \vz_{2}^{h}, \dots \vz_{F}^{h}])
\end{align}
Notably, $F$ is randomly chosen for each batch training to diversify the learning of different frequency wave activation.

\subsubsection{Frequency Contrastive Learning}
We introduce cross-band CL to unify the low-band and high-band representations.
This CL is also a dual prediction pretext.
The positive samples are $\vz_{F+K}^{l}$ and $\vz_{F+K}^{h}$ for $\vct^{l}$ and $\vct^{h}$, respectively.
The module employs linear projection $W^K_F(\cdot)$ following Transformer encoders to learn the frequency order information.

\begin{align}
L^l_{FC}:= \mathbb{E}_{\mathcal{X}}\adarecbracket{-\log \frac{\exp\adabracket{(W^K_F(\vct^{l})) \cdot \vz_{F+K}^l) / \tau}}{\sum_{n}\exp\adabracket{(W^K_F(\vct^{l}))\cdot \vz_{n}^l / \tau} }}, 
\\
L^h_{FC}:= \mathbb{E}_{\mathcal{X}}\adarecbracket{-\log \frac{\exp\adabracket{(W^K_F(\vct^{h})) \cdot \vz_{F+K}^h / \tau}}{\sum_{n}\exp\adabracket{(W^K_F(\vct^{h})) \cdot \vz_{n}^h / \tau} }}
\end{align}
where $\tau >0$ is a temperature parameter. 
Here, this module enhances the frequency representation quality by summarizing the transient waveforms (the high-frequency band) and long-term variations (the low-frequency band).

\subsection{Deep Clustering for Temporal-Frequency Alignment}\label{subsec:alignmodule}
We introduce a deep clustering learning to conduct an integration loss function, $L_\text{dc}$, for further optimization of temporal-frequency representation.
This loss function is an information metric with two expectations: 
(1) to align the two latent spaces.
ensuring their carried information is coherently related to \emph{the same input, i.e., same cluster assignment} and
(2) to ensure the learned representations are task-relevant by promoting \emph{distinct cluster assignments for other input samples}.
 
Formally, we initially concatenate representations as $\vct_T = \text{Concat}(\vct^{w}$, $\vct^{s})$ and $\vct_F = \text{Concat}(\vct^{l}$, $\vct^{h}$).
Inspired by SwAV \cite{SwAV}, we approach the clustering of $\vct_T$ and $\vct_F$ as a problem of similarity measures, aiming to maximize the mutual information between them.
Since $\vct_T$ and $\vct_F$ are from the same given $\vx$,
they inherently carry high informational similarity and conceptually belong to the same cluster.
We treat $\vct_T$ and $\vct_F$ as two distinct views of $\vx$, denoted $\vz^{v_1}$ and $\vz^{v_2}$, respectively. 
Therefore, high informational similarity between these views implies that they can predict each another.
The clustering task thus involves aligning a set of $J$ learnable cluster centroids $\{\vc_1, \dots, \vc_J\}$ to the paired views ($\vz^{v_1}, \vz^{v_2}$). 
\emph{If $\vz^{v_1}$ and $\vz^{v_2}$ can predict each other effectively, they are likely to share the same cluster centroids $\vc_j$.}
This process is given by:
\begin{align}
    L_\text{dc} := l(\vz^{v_1}, \vq^{v_2}) + l(\vz^{v_2}, \vq^{v_1})
\label{eq:swav}
\end{align}
where $\vq^{v_1}$, $\vq^{v_2}$ are trainable code that determines how to assign $\vz^{v_1}$ $\vz^{v_2}$, represented a mirror pair of a patient to a cluster centroid $\vc_j$.
In a nutshell, $l(\vz^{v_1}, \vq^{v_2})$ \emph{tells how close or aligned the latent variable $\vz^{v_1}$ is to all centroids, weighted by transformed vector $\vq^{v_2}$ of $\vz^{v_2}$}, given by:
\begin{align}
    l(\vz^{v_1}, \vq^{v_2}) &:= -\sum_{j}^{J} \vq_j^{v2} \log \frac{\exp\adabracket{\vz^{v_1}\cdot\vc_j / \tau}}{\sum_{j'} \exp\adabracket{\vz^{v_1}\cdot \vc_j' / \tau} }.
    \label{eq:swav}
\end{align}
The trainable code generalizes the Euclidean distance metric.
It can be seen that Eq.(\ref{eq:swav}) is a special case of InfoNCE in that the expectation is over the code.
Let $\vZ = [\vz_1,\dots,\vz_{b}]$ and $\vQ=[\vq_1,\dots,\vq_{b}]$ denote the matrices respectively collecting the representation vectors, where b refers to the batch size.
The code is trained by maximizing their similarity subject to the constraint of belonging to the transportation polytope \cite{SwAV}:
$\max_{Q} \text{Tr} \adabracket{ \vQ^T \vC^T \vZ} +  \mathcal{H}(\vQ)$.
$\mathcal{H}(\vQ)$  denotes the Shannon entropy of the code matrix in an element-wise manner to avoid deterministic solutions.
For each paired $\vct_T$ and $\vct_F$, this cluster assignment maximizes the similarity between the latent variables, ensuring the temporal and frequency representations are more related to the input. 

\subsection{Two-Phase Training}
\method is designed in combination with a two-phase learning strategy to enhance its practical applicability.

\noindent i) \textit{Pre-training phase} operates as an end-to-end training process. 
We conduct this phase in a fully self-supervised manner to obtain informative representations that are relevant to the data itself and can be transferred to various tasks.
The loss function is:
\begin{align}
\mathcal{L}_{Pretrain}= (L^w_{TC} +L^s_{TC} )+ (L^h_{FC}+L^l_{FC}) +L_{dc}
\end{align}

\noindent ii) \textit{Fine-tuning Phase}: 
After task-free pre-training, we implement task-oriented fine-tuning. 
Notably, the networks are partitioned, and all subsequent networks following the class token extraction are discarded. . 
This representation diverges from traditional approaches, as it is not simply the output of the features from a final layer in the model. 
Both temporal and frequency Transformers are pre-added class tokens, and we employ the intermediate class tokens $\vct ^t$ and $\vct ^f$, and concatenate them, denoted as 
$\vct := \text{Concat}(\vct ^t, \vct ^f)$. 
A task-specific training loss is applied with \textbf{only single linear classifier}.
We perform fine-tuning from the "input-to-intermediate-task" network on the representation, mapping it to the label information $\mathcal{Y}$, such as the classification task.
\section{Experimental Settings}
    \label{sec:exp}
    We assess the performance of \method to determine if it meets with the EXPECTATIONS (introduced in Section \ref{section:pre}). 
Our evaluation uses experiments to answer the following questions. 

\begin{itemize}
    \item [Q1.] {\it Effectiveness:}
        How effectively does it perform various downstream tasks with only single-channel EEG?
    \item [Q2.] {\it Robustness:}
        How well does \method learn meaningful representations without label supervision, and how adaptable is it to different channels?
    \item [Q3.] {\it Scalability:}
        How can knowledge be transferred across different datasets, and how can \method be adapted for real-world scenarios?
\end{itemize}

\subsection{Tasks and Datasets}
\begin{figure}[!t]
    \centering
    \begin{subfigure}[b]{0.23\textwidth}
        \includegraphics[width=\textwidth]{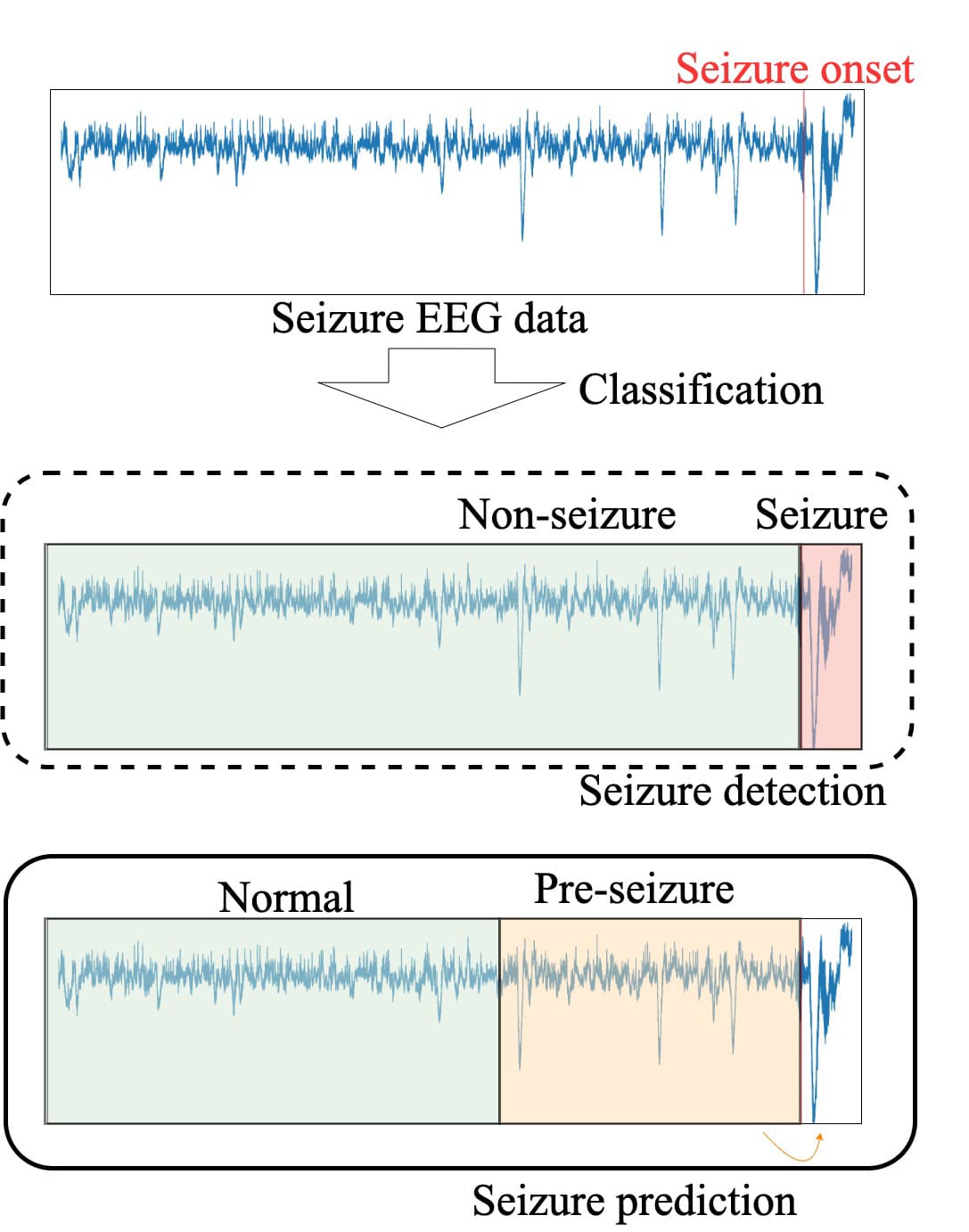}
        \caption{Seizure prediction task}
        \label{fig:seizure_description}
    \end{subfigure}
    \begin{subfigure}[b]{0.23\textwidth}
        \includegraphics[width=\textwidth]{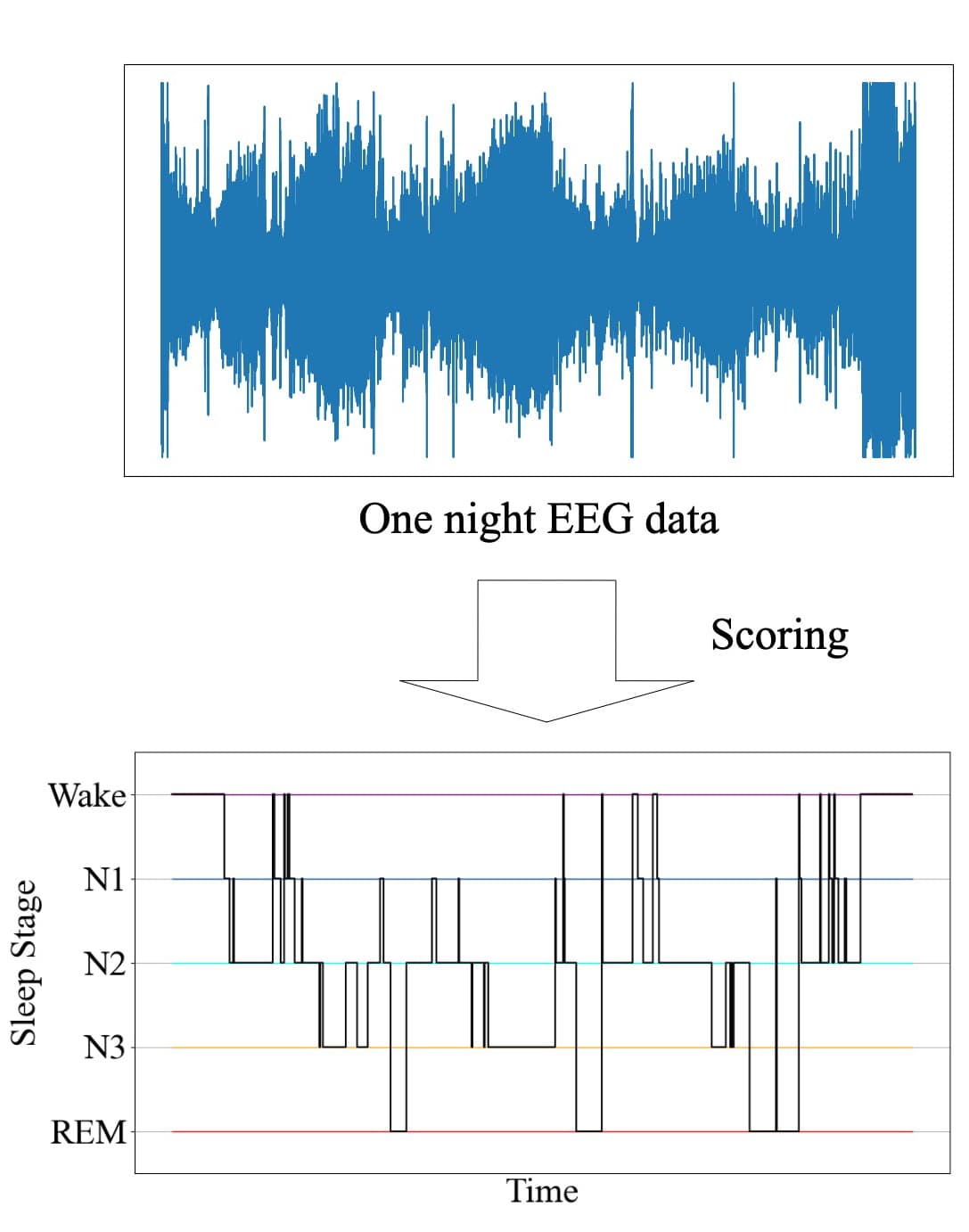}
        \caption{Sleep stage scoring task}
        \label{fig:sleep_description}
    \end{subfigure}
    \caption{Description of (a) seizure prediction and (b) sleep stage scoring tasks. (a) Seizure prediction task involves predicting epileptic seizures using EEG data designed to alert patients before an event. (b)Sleep stage scoring task categorizes sleep patterns into distinct stages, facilitating an understanding of sleep quality and disorders.}
    \label{fig:experiment_description}
\end{figure}

We conducted extensive experiments on four datasets with two EEG tasks: seizure prediction and sleep stage classification (a brief introduction can be found in Figure \ref{fig:experiment_description}).
The datasets are described in TABLE \ref{tab:dataset_summary}.

\noindent \textbf{Seizure prediction.} 
We focus on a more challenging task, that is, noticing early signs before the onset of a seizure. 
We used two public databases.
The CHB-MIT dataset\footnote{https://physionet.org/content/chbmit/} consists of 844 hours of 22-channel scalp EEG (sEEG) for 22 patients with 163 episodes of seizure. 
The pre-seizure state was considered to be 5 minutes before seizure onset.
The HUH dataset\footnote{https://zenodo.org/record/4940267\#.ZEVI3S9ByLd} consists of 21-channel sEEG of 79 patients. 
The pre-seizure state was considered to be 30 seconds before seizure onset.

\noindent \textbf{Sleep stage classification.} This task classifies sleep into five stages, i.e., wake, N1, N2, N3, and rapid eye movement (REM). 
We used two public databases, Sleep Heart Health Study (SHHS) and  Sleep-EDF.
SHHS\footnote{https://www.sleepdata.org/} is the largest available sleep dataset with 5793 patients.
The Sleep-EDF\footnote{https://www.physionet.org/content/sleep-edfx/1.0.0/} consists of 42308 sleep epochs from 20 healthy subjects.

\noindent \textbf{Our real clinical dataset.}
To verify the effectiveness in clinical diagnostics, we collected an Intracranial EEG (iEEG) dataset, namely OUH-iEEG, with epileptic seizure from Osaka University Hospital \cite{Kishima_ieeg}.
The dataset consists of iEEG from 13 patients, and the doctors selected the channel.  
The signals were acquired at 10000 Hz, downsampled to 2000 Hz, high-pass filtered at 0.5 Hz, and band-stop filtered at 60 Hz. 
The pre-seizure state was defined to be 20 seconds before seizure onset.

\subsection{Baselines}
We collected nine baselines for seizure prediction tasks.
\emph{(1) TS-TCC} \cite{TSTCC_ijcai21}: A self-supervised learning framework using temporal contrastive learning.
\emph{(2) TCN} \cite{TCN}: A neural network model that employs dilated convolutions to capture long-range temporal dependencies.
\emph{(3) MiniRocket} \cite{MiniRocket}: A convolutional kernel model based on the recent success of convolutional neural networks.
\emph{(4) STSF} \cite{STSF}: A method employing a tree-based structure and multiple time-series representations.
\emph{(5) WEASEL} \cite{WEASEL}: A method that transforms time series into feature vectors using FFT.
\emph{(6) Rasheed et al.} \cite{Rasheed}: A multi-channel seizure prediction method using a CNN.
\emph{(7) Chen et al.} \cite{Chen_SDM23}: A temporal-spatial contrastive learning framework to extract EEG features from a spectrogram.

For sleep stage scoring, we selected the following baselines: 
\emph{(1) TS-TCC} \cite{TSTCC_ijcai21}: A self-supervised learning framework using temporal contrastive learning.
\emph{(2) AttnSleep} \cite{AttnSleep_21}: A deep learning model employing attention mechanisms to capture long-range dependencies.
\emph{(3) DeepSleepNet-Lite} \cite{DeepSleepNet-Lite}: A CNN + RNN hybrid framework to model long and short temporal dependencies.
\emph{(4) Qu et al.} \cite{Qu_nn1}: A multi-scale deep architecture formed by decomposing an EEG signal into different frequency bands.
\emph{(5) SleepTransformer} \cite{SleepTransformer}: A deep learning model employing Transformer.
\emph{(6) SalientSleepNet} \cite{ijcai2021SalientSleepNet}: A framework composed of two-stream convolutional networks and multimodal attentions.
\emph{(7) Chen et al.} \cite{Chen_SDM23}: A temporal-spatial contrastive learning framework to extract EEG features from a spectrogram.
\emph{(8) Fernandez et al.} \cite{Enrique_Fernandez}: An ensemble of depth-wise separational convolutional neural networks.
\emph{(9) STSQ} \cite{STSQ}: A deep learning model using a CNN to extract spatio-temporal features from three modalities (EEG, EOG, and EMG), Bi-LSTM to extract sequential information.

\begin{table}[t]
\centering
\caption{Dataset description.}
\label{tab:dataset_summary}
\resizebox{\linewidth}{!}{%
\begin{tabular}{c|c|c|c|c|c}
\toprule
Task & \multicolumn{3}{c|}{Seizure prediction}& \multicolumn{2}{c}{Sleep stage classification} \\ 
\midrule
Dataset & CHB-MIT & HUH & OUH-iEEG & SleepEDF & SHHS \\
\midrule
\#Patient & 22 & 79 & 13 & 20 & 5793 \\
Data type & sEEG & sEEG & iEEG & sEEG & sEEG \\
\#Channel  & 22 & 21 & 1 & 1 (Fpz-Cz) & 1 (C4-A1) \\
Sampling Rate [Hz] & 256 & 256 & 2000 & 100 & 125 \\
Input size [second] & 2 & 1 & 1 & 30 & 30 \\
\bottomrule
\end{tabular}
}
\end{table}

\subsection{Evaluation Settings}
We used specificity (\emph{Spec.}), sensitivity (\emph{Sens.}), overall accuracy (\emph{Acc.}), and F1-score (\emph{F1}) to evaluate the performance.
Typically, EEG data are multi-channel data reflecting various brain regions. 
We conducted experiments for each channel individually and averaged the results.
We also evaluated the \emph{variance} across the different channels and their compatibility.
To demonstrate the efficacy of the self-supervised learning, we utilized uniform manifold approximation and projection (UMAP) to visualize the extracted features. 
We conducted these experiments using a server equipped with NVIDIA A6000 GPUs.
We tuned the hyper-parameters using Optuna \cite{optuna}, an optimization framework.
We used 20\% of the data for testing, 80\% for training, and within the training set, 20\% was designated as the validation set.

\begin{table*}[t]
    \centering
    \caption{Performance comparison on the CHB-MIT, HUH, and Our iEEG datasets.
    (-) denote that the evaluation was not applicable for multi-channel baselines, as the data set contained only one channel.
\textbf{Bold}: best, \underline{underscored}: second-best.
    }
    \label{tab:seizure_main}
    \resizebox{0.78\linewidth}{!}{%
    \begin{tabular}{lc|lll|ccc|ccc}
    \toprule
        & Datasets & \multicolumn{3}{c|}{CHB-MIT} & \multicolumn{3}{c|}{HUH} & \multicolumn{3}{c}{OUH-iEEG} \\ \midrule
         Baseline     & Channel & Spec. & Sens. & Acc. & Spec. & Sens. & Acc. & Spec. & Sens. & Acc. \\ \midrule
    Rasheed et al., 2021 \cite{Rasheed}      & Multi       & 0.890       & 0.909       & 0.920       & 0.692 & 0.675 & 0.687 & - & - & - \\
    Chen et al., 2023 \cite{Chen_SDM23} & Multi        & 0.930       & 0.926       & 0.932       & 0.835 & \underline{0.816} & \underline{0.821} & - & - & - \\
    \midrule
    WEASEL, 2017 \cite{WEASEL}      & single       & 0.921       & 0.906       & 0.914       & 0.762 & 0.681 & 0.721 & 0.806 & 0.685 & 0.756 \\
    TCN, 2018 \cite{TCN} & single        & 0.939       & 0.919       & 0.929       & 0.786 & 0.645 & 0.716 & 0.806 & 0.567 & 0.682 \\
    STSF, 2020 \cite{STSF}      & single       & 0.950       & \underline{0.927}       & \underline{0.939}       & \textbf{0.878} & 0.744 & 0.810 & 0.894 & 0.743 & 0.816 \\
    MiniRocket, 2021 \cite{MiniRocket}    & single         & 0.942       & 0.894       & 0.918       & 0.832 & 0.765 & 0.799 & \underline{0.916} & \underline{0.820} & \underline{0.866} \\
    TS-TCC, 2021 \cite{TSTCC_ijcai21} & single        & \underline{0.967}       & 0.875       & 0.922       & 0.624 & 0.778 & 0.701 & 0.833 & 0.706 & 0.767 \\
    \textbf{\emph{SplitSEE}}             & single        & \textbf{0.979}       & \textbf{0.984}       & \textbf{0.984}       & \underline{0.873} & \textbf{0.823} & \textbf{0.845} & \textbf{0.930} & \textbf{0.833} & \textbf{0.879} \\
    \bottomrule
    \end{tabular}
    }
\end{table*}

\begin{table}[t]
\centering
\caption{
Comparison on the SHHS and Sleep-EDF.
\textbf{Bold}: best, \underline{underscored}: second-best.
(*): single-channel baselines}
\label{tab:sleep_main}
\resizebox{\linewidth}{!}{%
\begin{tabular}{l|lllll|l}
\toprule
Baseline & Wake & N1 & N2 & N3 & REM & \emph{Acc.}\\
\midrule
\multicolumn{7}{c}{SHHS} \\
\midrule
Fernandez et al., 2021 \cite{Enrique_Fernandez} & 0.91 & 0.40 & 0.87 & 0.83 & 0.84 & 0.85 \\
STSQ, 2021 \cite{STSQ} & 0.92 & 0.40 & 0.84 & 0.76 & \textbf{0.89} & 0.84 \\
AttnSleep, 2021 \cite{AttnSleep_21} & 0.86* & 0.33* & \underline{0.87}* & 0.87* & 0.82* & 0.85* \\
Chen et al., 2023\cite{Chen_SDM23}& \underline{0.93} & 0.38 & \underline{0.88} & \underline{0.87} & 0.80 & 0.85 \\
SleepTransformer, 2021 \cite{SleepTransformer} & 0.92 & \textbf{0.46} & \underline{0.88} & 0.85 & \underline{0.88} & \textbf{0.88} \\
\textbf{\emph{SplitSEE}} & \textbf{0.94} & \underline{0.42} & \textbf{0.90} & \textbf{0.88} & 0.80 & \underline{0.86} \\
\midrule
\multicolumn{7}{c}{Sleep-EDF} \\
\midrule
Qu et al., 2020 \cite{Qu_nn1} & 0.90* & 0.48* & \underline{0.88}* & 0.86* & 0.83* & 0.84* \\
DeepSleepNet-Lite, 2021 \cite{DeepSleepNet-Lite} & 0.91* & 0.46* & 0.83* & 0.79* & 0.76* & \underline{0.85}* \\
TS-TCC, 2021 \cite{TSTCC_ijcai21} & 0.90* & 0.24* & \underline{0.88}* & \underline{0.88}* & 0.76* & 0.83* \\
SalientSleepNet, 2021 \cite{ijcai2021SalientSleepNet} & \underline{0.93} &\textbf{0.54} & 0.86 &0.78 & \textbf{0.86} & 0.84\\
SleepTransformer, 2021 \cite{SleepTransformer} & \underline{0.93} & \underline{0.49} & 0.87 & 0.81 & \underline{0.85} & \underline{0.85} \\
Chen et al., 2023\cite{Chen_SDM23} & 0.92 & 0.37 & 0.84 & 0.83 & 0.83 & 0.84 \\
\textbf{\emph{SplitSEE}} & \textbf{0.94} & 0.26 & \textbf{0.90} & \textbf{0.92} & 0.82 & \textbf{0.87} \\
\bottomrule
\end{tabular}
}
\end{table}

\section{Results}
    \label{sec:result}
    
\subsection{Effectiveness}

\mypara{Seizure prediction}
TABLE \ref{tab:seizure_main} shows the main results for the seizure prediction task performed on three datasets. 
Overall, our proposed \method outperforms all baselines on all metrics.
\method demonstrates superior performance compared with two multi-channel EEG baselines for the CHB-MIT dataset.
This promising performance can also be found in our real clinical datasets with 0.930, 0.833, and 0.879 of Spec, Sens, and Acc, respectively.
Our \method exhibits superior stability, with a notably small variance of 2.6 compared to the second-best method, with a variance of 19.2 for the CHB-MIT dataset.
In seizure prediction, accuracy is not the only critical metric. Sensitivity (correctly identifying seizures) and specificity (correctly ruling out non-seizures) are vital for providing timely alerts and minimizing false alarms.

\mypara{Sleep stage classification}
TABLE \ref{tab:sleep_main} shows that the proposed method achieved superior results to various single-channel EEG-only methods (marked by *) for SHHS datasets across all stage classifications.
Compared with the multi-data source baselines, \method also accurately classifies most of the classes, resulting in a high F1 with 0.94, 0.90, and 0.88 for Wake, N1, and N2, respectively.
A similar conclusion can be found in the results for the Sleep-EDF dataset.
The proposed method achieved higher staging performance even compared with other multi-channel methods \cite{ijcai2021SalientSleepNet,Chen_SDM23}.

\begin{figure}[t]
    \centering
\includegraphics[width=0.98\linewidth]{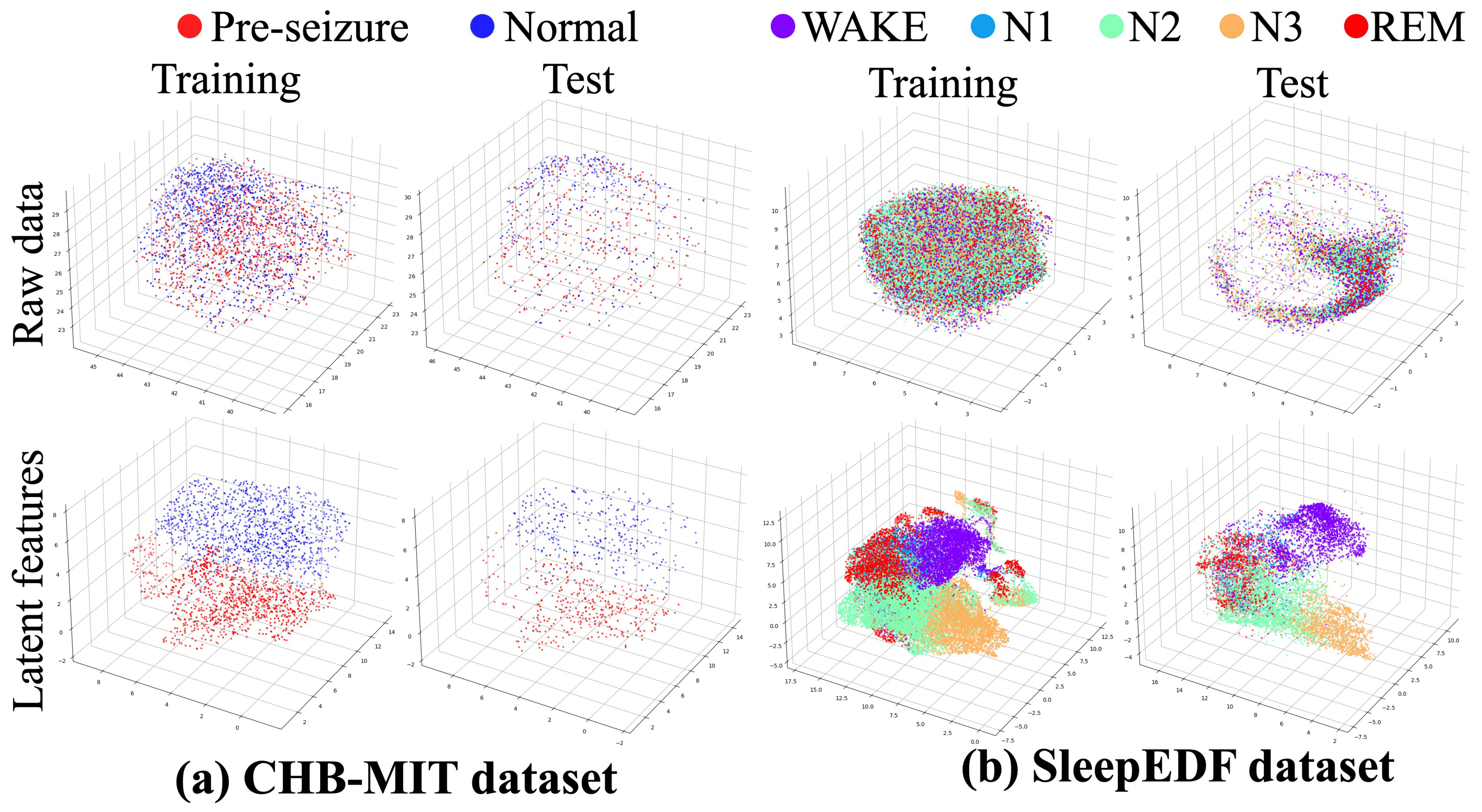}
    	    \vspace{-0.0em} \\
\vspace{-0.40em}
    \caption{UMAP visualization of raw data and features in (a) CHB-MIT and (b) SleepEDF datasets. The strength of the proposed method lies in the clear clustering boundaries observed in both training and test data for both tasks.}
    \vspace{-0.20em}
    \label{fig:visualization}
\end{figure}

\subsection{Robustness}

\mypara{Visualization}
Fig. \ref{fig:visualization} shows uniform manifold approximation and projection (UMAP) visualization of the features extracted by \emph{only self-supervised pre-training without any label supervision} in a 3D space. 
Despite the initial appearance of random dispersion in the raw data, there is a clear decision/clustering boundary of the self-supervised learning \method within the 3D space. 
We can achieve similarly robust representations with untrained test data by training on the training data.
A clear boundary can be found in two datasets, highlighting the high effectiveness of capturing an informative representation from different feature domains.
These results prove the power of our unsupervised learning pre-training.

\mypara{Variance}
Fig. \ref{fig:scatter} shows comparison results on the average accuracy among channels and its variance.
A smaller variance implies that the accuracies are closely grouped around the mean. 
Here, our method not only achieves the highest average accuracy but also shows the least variance.
This means that the errors and uncertainties in our method are minor, and we can achieve high classification accuracy no matter which channel we choose.
This result proves the effectiveness of extending \method to different EEG installations and settings.

\begin{figure}[t]
    \centering
    \includegraphics[width=\linewidth]{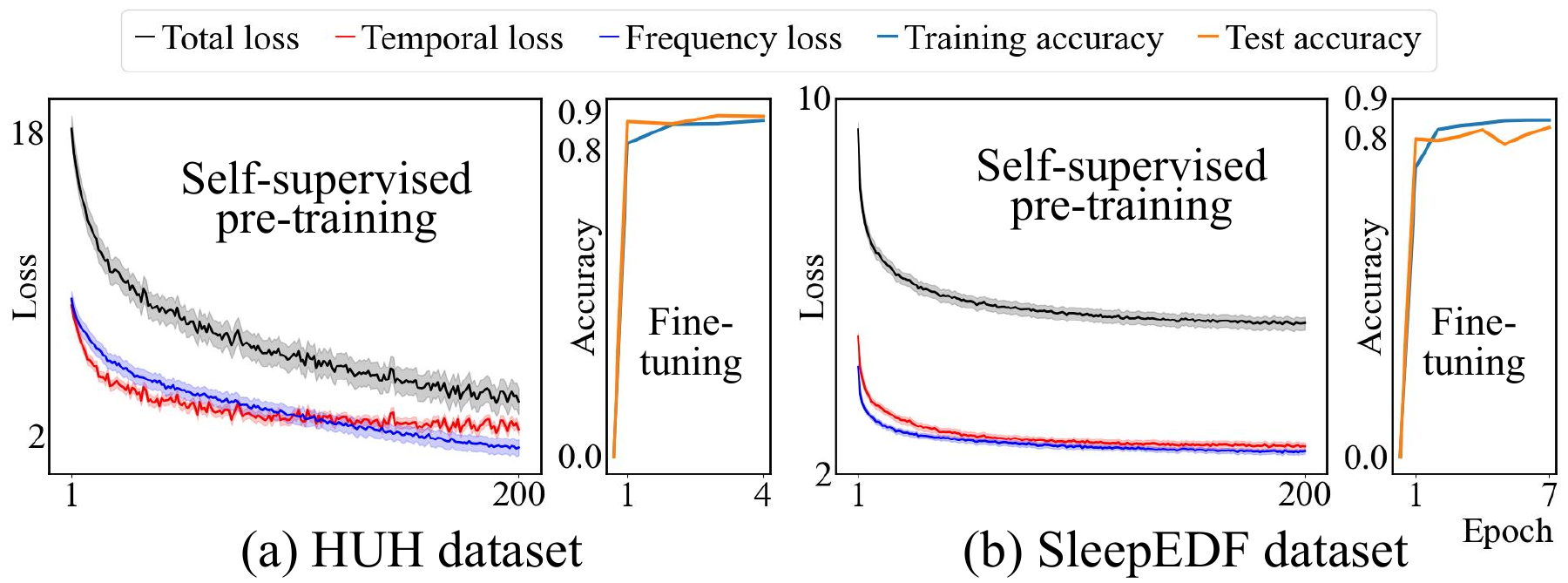}    \caption{Results of self-supervised training loss (left) and fine-tuned classification accuracy (right) on the (a) HUH and (b) SleepEDF datasets. Accuracy immediately rosed to around 80\% after the first fine-tuning epoch. 
    }
    \label{fig:curves}
\end{figure}

\subsection{Scalability}

\mypara{Loss and accuracy curves}
Fig. \ref{fig:curves} presents the self-supervised learning loss curve over 200 epochs alongside the fine-tuning accuracy curve for a few epochs. 
The loss curve demonstrates a consistent decline, highlighting the effectiveness of the loss function in guiding pre-training for both tasks. 
On the right, in Fig. \ref{fig:curves} (a) and (b), we also show results from supervised training on top of the self-learned features. 
Significant improvements are evident after just the first epoch of fine-tuning. 
Notably, this fine-tuning is performed only on selected networks, suggesting that superior performance can be achieved by using a few networks and minimal re-training. 
\section{Conclusion}
    \label{sec:conclusion}
    This paper presents \method, which is a splittable self-supervised learning framework that aims to extract representative and eventful features solely from single-channel EEG data.
Our method has the following desirable properties:
\begin{itemize}[left=0pt]
\item[(a)] { {Effectiveness}}:
\method learns informative features solely from single-channel EEG but has even outperformed multi-channel and multi-data source baselines.
\item[(b)]  { {Robustness}}: 
\method has the capacity to adapt across different channels with low performance variance. Superior performance is also achieved on four public databases and our real clinical dataset.
\item[(c)] {{Scalability}}:
Our experiments show that with just \emph{one fine-tuning epoch}, \method achieves high and stable performance using partial model layers.
The federated experiments mirror these results with only \emph{one-layer local deployment}, showing its great potential in real-world clinical scenarios.
\end{itemize}
Future directions include verifying the effectiveness of the proposed method in more fundamental neurological problems.

\mypara{Acknowledgment}
    {\small
The authors would like to thank the reviewers for their valuable comments and helpful suggestions.
This work was partly supported by
 JST BOOST, Japan Grant Number JPMJBS2402,
“Program for Leading Graduate Schools” of the Osaka University, Japan,
JSPS KAKENHI Grant-in-Aid for Scientific Research Number JP24K20778,
JST CREST JPMJCR23M3,
JST AIP Acceleration Research JPMJCR24U2,
AMED under Grant Number JP19dm0307103.
% The authors would like to thank the reviewers. %for their valuable comments and helpful suggestions.
}

% This research was supported by
% “Program for Leading Graduate Schools” of the Osaka University, Japan.

\bibliographystyle{IEEEtran}
\bibliography{main}

\end{document}